\documentclass[runningheads]{llncs}
\usepackage{graphicx}

\usepackage{float}
\usepackage{caption}
\usepackage{subcaption}
\usepackage{adjustbox}
\usepackage{tikz}
\usepackage{booktabs}
\usepackage{makecell}
\usepackage{comment,verbatim}
\usepackage{amsmath,amssymb} 
\usepackage{color}
\usepackage{booktabs}
\usepackage{todonotes}
\usepackage[accsupp]{axessibility}  

\begin{document}
\pagestyle{headings}
\mainmatter
\newcommand{\hugo}[1]{\todo[inline,color=green!40]{#1 -- Hugo}} 
\newcommand{\hugoc}[1]{\textcolor{blue!}{#1}}
\newcommand{\samethanks}[1][\value{footnote}]{\footnotemark[#1]}

\title{Towards benchmarking Western Bluebird detection in the wild} 

%
%
\institute{}


\titlerunning{Western Bluebird detection in the wild}
\author{Estela Monserrat Arriaga Santana\inst{1}\thanks{Equal contribution.} \and
Julian Rosas Scull\inst{1} \samethanks \and 
Ibeth P. Alarc\'on\inst{2} \and
Bibiana Montoya\inst{2} \and
Aylin Sosa Mej\'ia\inst{2} \and
Hugo Jair Escalante\inst{3}}
\authorrunning{E. M. Arriaga Santana et al.}
\institute{Universidad Nacional Aut\'onoma de M\'exico, Mexico\\
\email{\{monse\_arriaga,julian.rosas\}@ciencias.unam.mx}
\and
Universidad Aut\'onoma de Tlaxcala, Mexico\\
\email{ibethalarcon1@gmail.com, bibianac.montoyal@uatx.mx, aylinsm25@gmail.com}
\and
The University of Texas at El Paso, USA and INAOE, Mexico\\
\email{hescalantebal@utep.edu}
}
\maketitle

\begin{abstract}
Bird monitoring in natural environments is challenging due to the small size of some species of birds relative to the scene, background clutter, variability in illumination, and the observers' viewpoint. Progress is further limited by the scarcity of large-scale, realistic datasets, which are essential for understanding behavioral patterns. To address this gap, we introduce a new benchmark dataset for the detection and segmentation of Western bluebirds \textit{(Sialia Mexicana)}, comprising over 6,000 labeled images from 41 recording sessions. The dataset features high-resolution (4K) in-the-wild images in which birds occupy only a small fraction of the image. We evaluated a range of methods, including supervised detectors, 
open-vocabulary models 
under zero-shot and fine-tuned settings, and segmentation approaches. 
Our results show that the supervised detectors remain the most reliable overall, with Faster R-CNN achieving the highest detection mAP and RT-DETR offering the best precision–recall trade-off. Open-vocabulary models perform poorly in zero-shot settings; however, fine-tuning substantially improves their performance, with YOLO - World becoming competitive with supervised methods and achieving the highest precision, F1-score, and mAP@0.5. For segmentation, supervised methods significantly outperform Grounded-SAM and SAM 3: Mask R-CNN achieves the highest mask mAP, while YOLOv8-Seg provides the best precision and fastest inference. A diagnostic analysis further shows that failures are not explained by object size alone, but by a combination of apparent scale, brightness, contrast, clutter, blur, crowding, and recording-session variation. Overall, our findings highlight the difficulty of zero-shot bird detection in cluttered ecological scenes, suggest a more challenging regime for precise localization than is commonly observed in other wildlife and open-vocabulary benchmarks, and underscore the importance of domain adaptation in small-object settings.

\end{abstract}

%
%
\section{Introduction}

Computer vision has become increasingly relevant for ecological monitoring; however, bird detection in natural environments remains challenging due to small object size, background clutter, illumination variability, and frequent occlusion. This problem is particularly important, as insights into bird behavior can significantly advance our understanding of migration, breeding, and broader ecological patterns. In this context, we approach the problem of robust bird detection and localization in unconstrained environments.
\vspace{-1.2em}    
\begin{figure}[!htbp]
    \centering
    \includegraphics[width=0.3\linewidth]{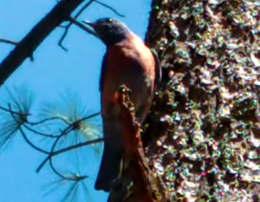}
    \caption{Close-up  of a male Western Bluebird, the species considered in this study.}
    \label{fig:bluebird_closeup}
\end{figure}
\vspace{-1.5em}

Specifically,  we consider the Western bluebird (\textit{Sialia mexicana}), a species approximately 19 cm in length, as a representative case study.
This species competes with both conspecific and heterospecific individuals for access to limited nesting sites, leading to rapid and frequent multi-individual interactions. These behaviors are difficult to accurately capture and annotate under natural conditions. As a result, video recordings are commonly used for behavioral analysis; however, their manual inspection requires substantial effort from human observers to identify and classify interactions.

Our work aims to advance the understanding of the behavior of this species. In particular, we expand a novel video dataset of Western bluebirds, 
originally acquired for manual analysis in the context of behavioral ecology research~\cite{alarcon2025efectos}, to enable the systematic evaluation of computer vision methods for monitoring behavioral interactions in natural conditions.
Source recordings were collected during a field experiment conducted during the 2024 breeding season on nest-site competition in wild individuals of \textit{Sialia mexicana}, breeding in nest boxes under natural conditions.

Unlike standard bird datasets, which often focus on close-up classification or aerial monitoring, ours targets a fixed-camera behavioral ecology-study scenario in which birds occupy a small fraction of high-resolution frames, interact with multiple individuals, and are subject to real-world variability such as occlusion and illumination changes, see Figure~\ref{fig:dataset_examples}.
This setup enables a systematic comparison of different detection paradigms under realistic ecological conditions. We evaluate representative state-of-the-art models across supervised, transformer-based, and open-vocabulary paradigms.
Our contributions are threefold:
\begin{itemize}
    \item We present a curated and manually annotated dataset derived from ecological video recordings of bird behavior.
    \item We provide a comparative benchmark of supervised, transformer-based, and open-vocabulary models for bird detection, with segmentation as a complementary analysis. 
    \item We show that fixed-camera bird detection under natural conditions remains a challenging small-object regime in which supervised adaptation is still more reliable than zero-shot open-vocabulary detection.
\end{itemize}

\section{Related Work}

Camera-trap and ecological imaging datasets have enabled large-scale wildlife monitoring and automated analysis~\cite{swanson2015snapshot,norouzzadeh2018automatically}. However, models trained in one ecological setting often struggle to generalize across regions and environments due to changes in species composition, illumination, and background clutter \cite{beery2019efficient}. In bird monitoring, existing datasets are often based on aerial imagery~\cite{weinstein2022general,hayes2021drones,hong2023bird} or fine-grained classification benchmarks such as CUB-200-2011, NABirds, and Birdsnap \cite{wah2011cub,van2015building,berg2014birdsnap}, where birds appear large and centered in the frame. By contrast, our dataset focuses on fixed-camera bird detection under challenging natural conditions, where birds occupy only a small fraction of high-resolution images and appear with clutter, motion blur, and viewpoint variation.

Object detection has evolved from two-stage methods such as Faster R-CNN \cite{ren2015faster} to one-stage detectors such as YOLO \cite{redmon2016yolo}, and more recently to transformer-based approaches including DETR and RT-DETR \cite{carion2020end,zhao2024rtdetr}. Open-vocabulary detectors such as OWL-ViT and YOLO-World \cite{minderer2022simple,cheng2024yoloworld}, as well as grounding-based models like Grounding DINO \cite{liu2023grounding}, extend detection beyond fixed label sets through vision-language alignment. In parallel, segmentation models such as SAM~\cite{kirillov2023segment} enable prompt-based mask generation. Our benchmark builds on these directions by comparing supervised, transformer-based, and open-vocabulary methods in a fixed-camera setting.

\section{A dataset for monitoring Western bluebird in the wild}
The source dataset was collected in a controlled behavioral experiment conducted at the \textbf{La Malinche National Park (Tlaxcala, Mexico)} by staff of \textbf{Estación Científica La Malinche}, part of the \emph{Centro Tlaxcala de Biología de la Conducta, Universidad Autónoma de Tlaxcala}. Recordings captured the activity of 33 breeding pairs of Western bluebirds that interacted with a potential competitor for the nesting site, from  a different species (Chipping sparrow o House finch), which was placed inside a wire mesh box, in a forested area. While this setup is controlled in terms of camera placement and viewpoint, bird behavior occurs naturally, resulting in realistic ecological variability~\cite{alarcon2025efectos}.
\subsection{Source videos}
\begin{figure} [h!t]
    \centering
    \includegraphics[width=0.6\linewidth]{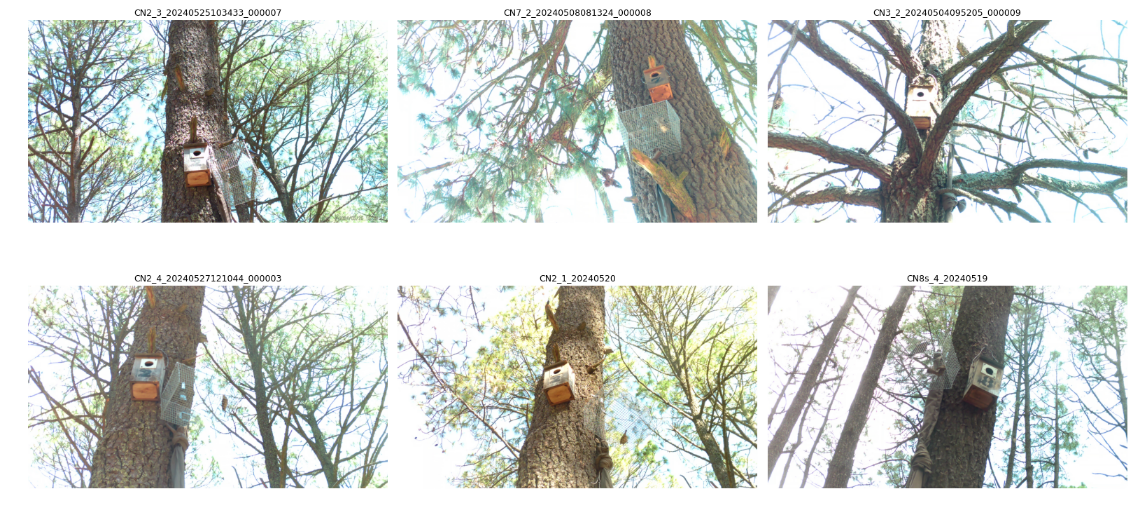}
    \caption{Sample images from the dataset.
    }
    \label{fig:dataset_examples}
\end{figure}

Source videos were recorded using fixed-position cameras directed toward nest boxes, ensuring a consistent field of view within each recording, see~\cite{alarcon2025efectos} for details. Figure~\ref{fig:dataset_examples} shows sampled frames from the collection. All videos were captured at high resolution ($3840 \times 2160$). Each experimental trial lasted approximately 13 minutes and was repeated 4 or 5 times per nest. Only the first 7 minutes of each of the 41 recordings were used, corresponding to the period immediately after a potential intruder from another species, placed in a wire mesh box, is uncovered, when bird activity is highest. Cameras were placed at different orientations across sessions, resulting in variations in viewpoint, scale, and background composition (Figure~\ref{fig:dataset_examples}).
\vspace{-2.5em}

\subsection{Dataset and Annotations}

Frames were extracted from the recorded videos and manually annotated using LabelMe~\cite{russell2008labelme}.
Each image contains bounding box annotations for individual bird instances, and a subset includes instance segmentation masks for segmentation experiments.

The first version of the dataset consists of \textbf{6,016 images} and \textbf{7,637 annotated bird instances}, with an average of \textbf{1.27 instances per image}. The number of birds per image ranges from one to four: 4,694 with one, 1,043 with two, 259 with three, and 20 with four.

All images are in high-resolution ($3840 \times 2160$), but birds occupy a very small portion of the frame. The median bounding box area is approximately $2.6 \times 10^4$ pixels (less than $0.3\%$ of the image), making the dataset one of the most challenging for small-object detection.

\begin{figure}[h]
\centering
\includegraphics[width=0.6\linewidth]{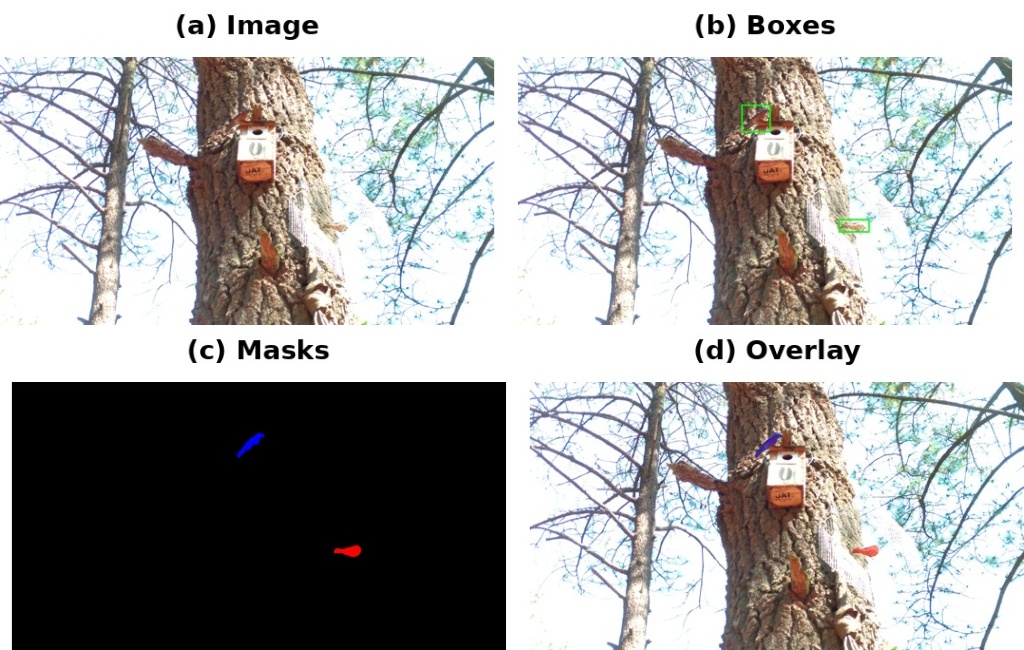}
\caption{
Example from the proposed bird detection dataset.
(a) Original image captured under challenging real-world conditions.
(b) Ground-truth bounding box annotations.
(c) Instance-level segmentation masks.
(d) Combined visualization of annotations.
}
\label{fig:annotation_examples}
\end{figure}

The dataset is organized into 41 independent recording sessions, each corresponding to a distinct experimental trial. Due to the recording setup, all images contain at least one bird instance. Figure~\ref{fig:annotation_examples} shows representative annotated examples.

\section{Methodology}

We benchmark bird detection in fixed-camera recordings under a unified experimental setup, with segmentation included as a secondary analysis. We evaluate supervised detectors (YOLOv8, YOLO26, Faster R-CNN, and RT-DETR) and open-vocabulary models (OWL-ViT, Grounding DINO, and YOLO-World). Supervised models are fine-tuned on the training split using COCO-pretrained weights \cite{lin2014coco}. Open-vocabulary models are evaluated in zero-shot (ZS) mode using a set of semantically related text prompts, including \texttt{bird}, \texttt{a bird}, \texttt{a flying bird}, \texttt{a perched bird}, and \texttt{bird in a cage}. When supported, models are also evaluated in a fine-tuned (FT) setting. OWL-ViT and YOLO-World are evaluated in both ZS and FT regimes, whereas Grounding DINO is evaluated only in the zero-shot setting.

Fine-tuned models are trained for up to 100 epochs with early stopping based on validation performance (patience = 20), providing a consistent training budget across methods. For each input image, models predict bounding boxes with confidence scores, and open-vocabulary methods additionally condition predictions on the text query. To complement detection, we evaluate segmentation using two supervised baselines, YOLOv8-Seg and Mask R-CNN, together with Grounded-SAM as a prompt-based zero-shot pipeline that combines Grounding DINO \cite{liu2023grounding} with SAM \cite{kirillov2023segment}, and SAM 3 \cite{carion2025sam3segmentconcepts} as an open-vocabulary zero-shot model.

We report standard COCO-style metrics. For detection, we use mAP@0.5:0.95, mAP@0.5, Precision, Recall, F1-score, and FPS. For segmentation, we report mask mAP@0.5:0.95, mask mAP@0.5, Precision, Recall, F1-score, mIoU, Dice, and FPS. This setup enables a consistent comparison of supervised, transformer-based, and open-vocabulary methods in a challenging small-object ecological setting.

\begin{figure}[h]
    \centering
    \includegraphics[width=0.7\linewidth]{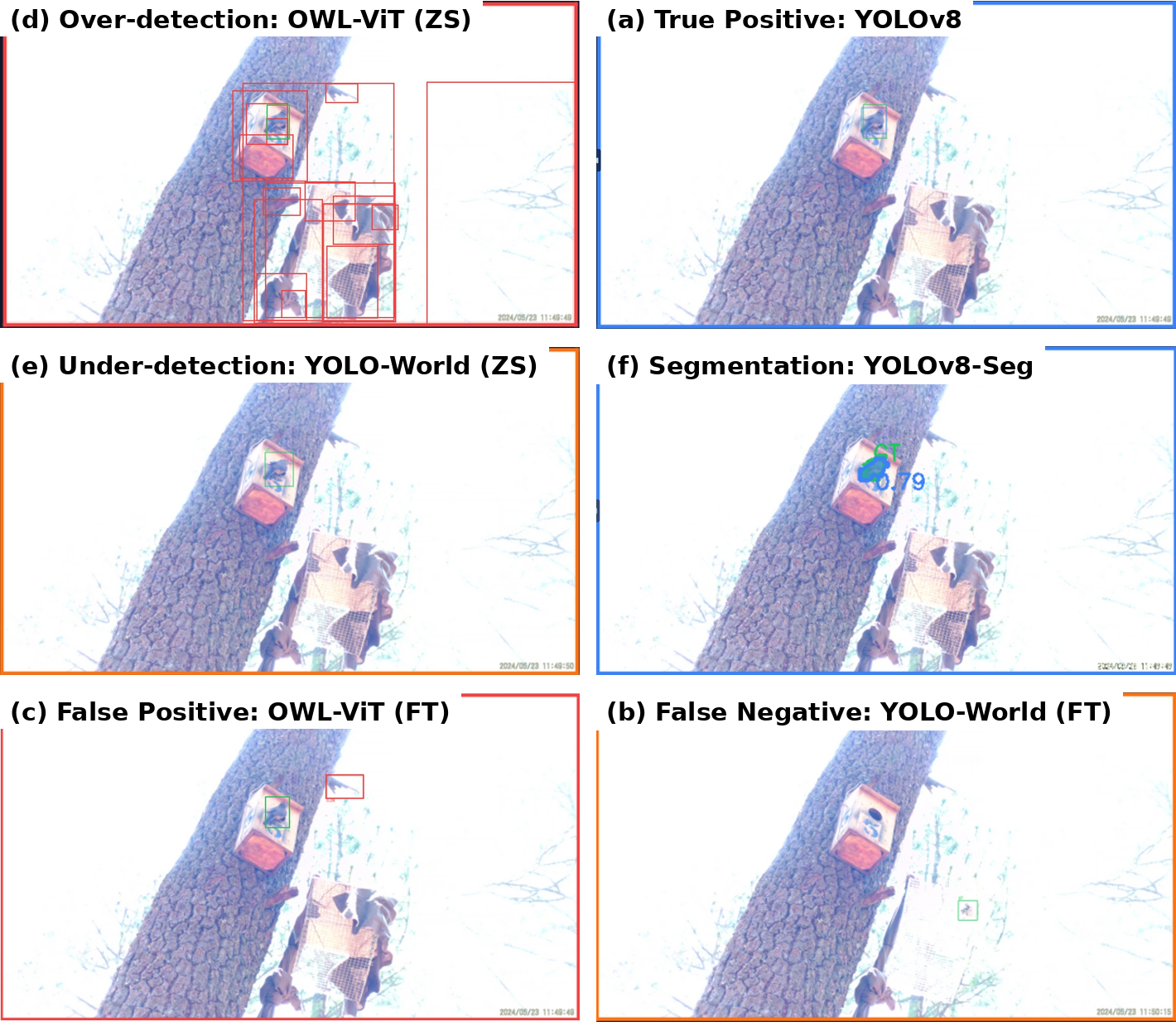}
    \caption{
Qualitative results on the test set.
(a) True positive from YOLOv8.
(b) False negative from YOLO-World (FT).
(c) False positive from OWL-ViT (FT).
(d) Over-detection from OWL-ViT (ZS).
(e) Under-detection from YOLO-World (ZS).
(f) Successful segmentation from YOLOv8-Seg.
Ground truth is shown in green, correct predictions in blue, and incorrect predictions in red.
}
    \label{fig:qualitative_results}
\end{figure}
\vspace{-1.5em}

\section{Experiments and results}

\subsection{Experimental Setup}

The dataset is split at the recording level to avoid temporal leakage across training, validation, and test sets. In total, it contains 41 recording sessions and 6,016 annotated images, divided into 35 training videos (4,435 images), 5 validation videos (575 images), and 1 held-out test video (1,006 images). Validation set is used for model selection and early stopping, final results are reported on the unseen test split.

All experiments were conducted on the Kaggle platform using NVIDIA T4 GPUs. When
supported by the implementation, training was performed with two T4 GPUs;
otherwise, a single T4 GPU was used. For each model family, the input-resolution
setting was kept fixed between training and inference to ensure a consistent
evaluation protocol within each implementation.

\vspace{-1em}
\subsection{Detection Results}

Table~\ref{tab:detection_results} summarizes detection performance on the test set. Among supervised detectors, Faster R-CNN achieves the highest mAP@0.5:0.95, indicating superior localization under stricter IoU thresholds, while RT-DETR provides the best overall balance with the highest recall and strong precision and F1-score, as well as the highest mAP@0.5 among supervised methods. YOLO-based models offer an efficient trade-off: YOLOv8 performs competitively, while YOLO26 improves precision and recall, achieving the highest F1-score and fastest inference speed, although with lower mAP@0.5:0.95 than Faster R-CNN and RT-DETR. Open-vocabulary models are unstable in the zero-shot setting: OWL-ViT over-detects, YOLO-World under-detects, and Grounding DINO remains below supervised baselines. After fine-tuning, YOLO-World becomes competitive, achieving the highest mAP@0.5, precision, and F1-score, though still below Faster R-CNN in mAP@0.5:0.95, while OWL-ViT improves but remains weaker. Overall, recent real-time detectors such as YOLO26 improve efficiency, but supervised models remain more reliable for precise localization, and domain adaptation is key for open-vocabulary methods.

\begin{table}[H]
\centering
\setlength{\tabcolsep}{3pt}
\renewcommand{\arraystretch}{1.08}
\begin{tabular}{llcccccc}
\toprule
\textbf{Paradigm} & \textbf{Model} & \textbf{mAP} & \textbf{mAP@0.5} & 
\textbf{Precision} & \textbf{Recall} & \textbf{F1} & \textbf{FPS} \\
\midrule
Sup. / FT & YOLOv8        & 0.3804 & 0.8428 & 0.8977 & 0.7338 & 0.8076 & 13.49 \\
Sup. / FT & YOLO26       & 0.3633 & 0.8772 & 0.9704 & 0.8852 & 0.9259 & \textbf{13.78} \\
Sup. / FT & Faster R-CNN  & \textbf{0.4662} & 0.9026 & 0.4500 & 0.8987 & 0.5997 & 7.36 \\
Sup. / FT & RT-DETR       & 0.4170 & 0.9340 & 0.9120 & \textbf{0.9210} & 0.9170 & 5.50 \\
Op-v. / ZS & OWL-ViT      & 0.0070 & 0.0189 & 0.0376 & 0.4609 & 0.0695 & 3.34 \\
Op-v. / FT & OWL-ViT      & 0.1264 & 0.2770 & 0.2002 & 0.2546 & 0.2241 & 4.18 \\
Op-v. / ZS & Grounding DINO & 0.0230 & 0.0880 & 0.0880 & 0.0920 & 0.0900 & 1.18 \\
Op-v. / ZS & YOLO-World   & 0.1560 & 0.4540 & 0.9000 & 0.0080 & 0.0170 & 12.37 \\
Op-v. / FT & YOLO-World   & 0.4467 & \textbf{0.9415} & \textbf{0.9810} & 0.8968 & \textbf{0.9370} & 5.19 \\
\bottomrule
\end{tabular}
\vspace{0.4em}
\caption{Detection benchmark results on the test set. mAP denotes box AP.}
\label{tab:detection_results}
\end{table}
\vspace{-4em}

\subsection{Segmentation Results}

Table~\ref{tab:segmentation_results} reports segmentation performance on the test set. Both supervised methods substantially outperform the prompt-based Grounded-SAM baseline, which fails completely in the zero-shot setting. Mask R-CNN achieves the highest mask mAP, recall, mIoU, and Dice, indicating the strongest overall segmentation quality.

\begin{table}[H]
\centering
\setlength{\tabcolsep}{2pt}
\renewcommand{\arraystretch}{1.08}
\begin{tabular}{llccccccccc}
\toprule
\textbf{Paradigm} & \textbf{Model} & \textbf{mAP} & \textbf{mAP@0.5} & 
\textbf{Precision} & \textbf{Recall} & \textbf{F1} & \textbf{mIoU} & \textbf{Dice} & \textbf{FPS} \\
\midrule
Sup. / FT & YOLOv8-Seg    & 0.1820 & 0.6653 & \textbf{0.7991} & 0.7136 & 0.7539 & 0.6459 & 0.7823 & \textbf{13.83} \\
Sup. / FT & Mask R-CNN    & \textbf{0.6398} & \textbf{0.9154} & 0.6701 & \textbf{0.9441} & \textbf{0.7838} & \textbf{0.8488} & \textbf{0.9166} & 7.75 \\
Prmpt. / ZS & Grounded-SAM & 0.0000 & 0.0000 & 0.0000 & 0.0000 & 0.0000 & 0.0000 & 0.0000 & 0.91 \\
Op-v. / ZS & SAM 3        & 0.2229 & 0.3186 & 0.2282 & 0.7252 & 0.3472 & 0.1793 & 0.1931 & 0.29 \\
\bottomrule
\end{tabular}
\vspace{0.4em}
\caption{Segmentation benchmark results on the test set. mAP denotes mask AP.}
\label{tab:segmentation_results}
\end{table}
\vspace{-3em}

In contrast, YOLOv8-Seg provides the highest precision and the fastest inference, showing a favorable trade-off between localization quality and efficiency. 
Although segmentation is not the primary task of our benchmark, these results suggest that supervised pixel-level localization can be practically advantageous for bird identification in cluttered natural scenes, especially when accurate spatial delineation is required. More broadly, the large gap between the supervised methods and Grounded-SAM highlights that prompt-based zero-shot segmentation is not yet reliable for this small-object ecological scenario.

\section{Qualitative analysis}

To complement the aggregate benchmark results, we conduct a dedicated diagnostic analysis on a separate error-analysis set, distinct from the test split used in Section 5, examining how the performance of representative models varies under object-scale, visual, and scene-level difficulty factors that commonly arise in field-based bird monitoring. We analyze models spanning detection, segmentation, and supervision regimes: RT-DETR FT, YOLO-World ZS/FT, Mask R-CNN FT, and SAM 3 ZS. The error-analysis set comprises 505 additionally annotated 4K frames with 529 bird instances, sampled from four new recording sessions. For the evaluation of object-scale influence, we report AP under two object-size definitions: COCO-style absolute bins, based on bounding-box area in pixels~\cite{lin2014coco}, and USB-style relative-scale bins, based on the fraction of the image occupied by the object~\cite{shinya2021usb}. For assessing visual difficulty factors, we use greedy recall instead of AP, matching predictions to ground-truth instances at IoU $\geq 0.5$ with a confidence threshold of 0.25, because this directly measures whether annotated bird instances are recovered or missed under each condition. Continuous factors are discretized using quantile thresholds computed on the error-analysis set, whereas count-based factors are grouped according to their natural categories.

Occlusion and distance to image borders were considered but excluded from the model-specific discussion due to insufficient variation: instances were effectively non-occluded, and near-border cases were too scarce for reliable stratified analysis.

\subsection{Scale-aware object-size analysis}

Bird instances occupy only a small fraction of the 4K frames, making object scale
a relevant factor in our dataset. We therefore evaluate size using both
COCO-style absolute scale bins and relative-scale bins based on apparent object
size. COCO scale is computed from absolute bounding-box area, but this definition
is only partially informative in our setting: AP$_S$ is undefined because no
instances fall into the COCO \textit{small} category, even though the birds are
visually small relative to the full image.

To capture this apparent scale, we use the relative scale index
\begin{equation}
s_r = \sqrt{\frac{wh}{WH}},
\label{eq:relative_scale}
\end{equation}
where \(w,h\) are the bounding-box dimensions and \(W,H\) are the image dimensions. For the AP-based analysis, we use fixed USB-style relative-scale bins~\cite{shinya2021usb}. In the error-analysis split, only two bins contain ground-truth instances: \(1/32\text{--}1/16\) with 350 instances and \(1/16\text{--}1/8\) with 179 instances.

Table~\ref{tab:scale_analysis_compact} shows that the fine-tuned detectors and segmentation models remain relatively stable across the two populated relative-scale bins, suggesting that object size alone does not explain their failures. In contrast, YOLO-World ZS shows the clearest scale-dependent behavior, with AP increasing from 0.1148 to 0.2050 as relative object size increases. This suggests that apparent scale affects zero-shot detection more strongly, while failures in the fine-tuned and segmentation models depend on a broader combination of visual and scene-level factors.

\begin{table}[H]
\centering
\small
\setlength{\tabcolsep}{5pt}
\renewcommand{\arraystretch}{1.12}
\begin{tabular}{lcccc}
\toprule
\textbf{Model} & 
\textbf{COCO-M} & 
\textbf{COCO-L} & 
\(\mathbf{1/32\text{--}1/16}\) & 
\(\mathbf{1/16\text{--}1/8}\) \\
\midrule
YOLO-World FT  & 0.5692 & 0.5311 & 0.5206 & 0.5329 \\
RT-DETR FT     & 0.5028 & 0.4304 & 0.4404 & 0.4313 \\
YOLO-World ZS  & 0.0436 & 0.1902 & 0.1148 & 0.2050 \\
Mask R-CNN FT  & 0.7223 & 0.7305 & 0.7363 & 0.6894 \\
SAM 3 ZS        & 0.7373 & 0.7504 & 0.7341 & 0.7159 \\
\bottomrule
\end{tabular}

\vspace{0.6em}
\caption{AP-based scale analysis on the error-analysis split. COCO-M and COCO-L report COCO-style AP, while the last two columns report AP over the only populated USB-style relative-scale bins.}
\label{tab:scale_analysis_compact}
\end{table}
\vspace{-4em}

\subsection{Visual difficulty-factor analysis}

\subsubsection{Difficulty-factor definitions}\mbox{}\\[0.2em]
Tables~\ref{tab:brightness_difficulty}--\ref{tab:crowding_difficulty} summarize the recall-based difficulty analysis by factor type. Each entry reports the worst-performing bin for a given model and factor, followed by greedy recall and the recall drop $\Delta$ with respect to the best-performing bin of the same factor.

\leavevmode\par\vspace{0.3em}
\textbf{Brightness} is computed as the mean grayscale intensity. We consider both global brightness, measured over the full frame, and local brightness, measured inside the ground-truth bird bounding box. We refer to this factor as brightness rather than physical illumination, since it measures the average gray value observed in the image. This follows image-quality assessment practice in which the average gray value is used to represent image brightness~\cite{li2022brightness}.

\vspace{-0.4em}
\begin{table}[H]
\centering
\scriptsize
\setlength{\tabcolsep}{4pt}
\renewcommand{\arraystretch}{1.12}
\begin{tabular}{lcc}
\toprule
\textbf{Model} & \textbf{Global brightness} & \textbf{Local brightness} \\
\midrule
YOLO-World FT & Neutral: 85.6\%, $\Delta$=12.1 & Dark: 77.7\%, $\Delta$=20.6 \\
RT-DETR FT & Dark: 84.0\%, $\Delta$=15.4 & Dark: 77.7\%, $\Delta$=20.6 \\
Mask R-CNN FT & Neutral: 93.7\%, $\Delta$=5.2 & Bright: 95.6\%, $\Delta$=2.7 \\
YOLO-World ZS & Bright: 22.8\%, $\Delta$=24.7 & Dark: 19.4\%, $\Delta$=26.0 \\
SAM 3 ZS & Neutral: 83.9\%, $\Delta$=11.6 & Dark: 86.3\%, $\Delta$=8.0 \\
\bottomrule
\end{tabular}
\vspace{0.4em}
\caption{Brightness difficulty analysis.}
\label{tab:brightness_difficulty}
\end{table}
\vspace{-3em}

\textbf{Contrast} is measured using Root Mean Square (RMS) contrast, computed as the standard deviation of grayscale pixel intensities. As with brightness, we compute both a global version over the full frame and a local version inside the ground-truth bird bounding box. RMS contrast is defined as the root mean square deviation of gray-level values from their mean, which is equivalent to the population standard deviation of grayscale intensities up to the chosen intensity
scale~\cite{vitor2023rmscontrast}.

\vspace{-0.4em}
\begin{table}[H]
\centering
\scriptsize
\setlength{\tabcolsep}{4pt}
\renewcommand{\arraystretch}{1.12}
\begin{tabular}{lcc}
\toprule
\textbf{Model} & \textbf{Global contrast} & \textbf{Local contrast} \\
\midrule
YOLO-World FT & Medium: 86.2\%, $\Delta$=8.7 & Low: 86.9\%, $\Delta$=7.0 \\
RT-DETR FT & Medium: 84.5\%, $\Delta$=10.9 & Low: 85.7\%, $\Delta$=9.3 \\
Mask R-CNN FT & Low: 94.3\%, $\Delta$=4.0 & Low: 93.7\%, $\Delta$=5.7 \\
YOLO-World ZS & High: 21.7\%, $\Delta$=20.0 & Low: 17.7\%, $\Delta$=38.4 \\
SAM 3 ZS & Low: 87.4\%, $\Delta$=4.2 & Low: 81.1\%, $\Delta$=13.9 \\
\bottomrule
\end{tabular}
\vspace{0.4em}
\caption{Contrast difficulty analysis.}
\label{tab:contrast_difficulty}
\end{table}
\vspace{-3em}

\textbf{Blur} is assessed through Laplacian-variance sharpness, computed as the variance of the Laplacian response over the grayscale bird crop. This follows classical focus-measure approaches based on image gradients or Laplacian responses, where sharp images produce stronger high-frequency responses and blurred images yield smoother, lower-variance responses~\cite{pechpacheco2000diatom}. Therefore, in this analysis, lower Laplacian-variance values correspond to blurrier bird crops, whereas higher values indicate sharper local boundaries.

\vspace{-0.4em}
\begin{table}[H]
\centering
\scriptsize
\setlength{\tabcolsep}{4pt}
\renewcommand{\arraystretch}{1.12}
\begin{tabular}{lc}
\toprule
\textbf{Model} & \textbf{Local blur / sharpness} \\
\midrule
YOLO-World FT & Medium: 87.4\%, $\Delta$=7.1 \\
RT-DETR FT & Medium: 85.1\%, $\Delta$=11.6 \\
Mask R-CNN FT & Blurred: 93.7\%, $\Delta$=6.3 \\
YOLO-World ZS & Blurred: 20.0\%, $\Delta$=30.6 \\
SAM 3 ZS & Blurred: 84.6\%, $\Delta$=10.4 \\
\bottomrule
\end{tabular}
\vspace{0.4em}
\caption{Blur/sharpness difficulty analysis.}
\label{tab:blur_difficulty}
\end{table}
\vspace{-3em}

\textbf{Relative size} is measured using Eq.~\ref{eq:relative_scale}. For the
difficulty-factor analysis, instances are grouped into quartile-based
relative-size bins and evaluated with greedy recall. This tests whether birds that occupy a smaller fraction of the 4K frame are more likely to be missed.

\vspace{-0.4em}
\begin{table}[H]
\centering
\scriptsize
\setlength{\tabcolsep}{4pt}
\renewcommand{\arraystretch}{1.12}
\begin{tabular}{lc}
\toprule
\textbf{Model} & \textbf{Relative size $s_r$} \\
\midrule
YOLO-World FT & Very small: 78.3\%, $\Delta$=21.1 \\
RT-DETR FT & Very small: 83.5\%, $\Delta$=15.2 \\
Mask R-CNN FT & Large: 93.2\%, $\Delta$=6.1 \\
YOLO-World ZS & Very small: 17.4\%, $\Delta$=33.4 \\
SAM 3 ZS & Very small: 85.2\%, $\Delta$=10.8 \\
\bottomrule
\end{tabular}
\vspace{0.4em}
\caption{Object-size difficulty analysis using the relative scale index
$s_r$.}
\label{tab:size_difficulty}
\end{table}
\vspace{-2.5em}

\textbf{Background clutter.}
We approximate background clutter using Canny edge density. For each frame, a binary Canny edge map is computed, and ground-truth bird boxes are excluded before measuring the fraction of edge pixels in the remaining background. This adapts edge-density visual-complexity measures to object-detection failure analysis: edge density is defined as the proportion of edge pixels in an image, and Canny edges provide the binary edge map used in the numerator \cite{chu2025visualcomplexity,canny1986computational}. Excluding the annotated bird boxes prevents the bird boundaries from inflating the clutter estimate.

Formally, background clutter is computed as
\begin{equation}
C_{\mathrm{bg}} =
\frac{
\sum_{p \in \Omega} E(p) M_{\mathrm{bg}}(p)
}{
\sum_{p \in \Omega} M_{\mathrm{bg}}(p)
},
\end{equation}
where \(E(p)\) is the binary Canny edge map and \(M_{\mathrm{bg}}(p)\) is a
binary mask that excludes ground-truth bird boxes.

\vspace{-0.4em}
\begin{table}[H]
\centering
\scriptsize
\setlength{\tabcolsep}{4pt}
\renewcommand{\arraystretch}{1.12}
\begin{tabular}{lcc}
\toprule
\textbf{Model} & \textbf{Background clutter} & \textbf{Local clutter} \\
\midrule
YOLO-World FT & Cluttered: 79.4\%, $\Delta$=18.3 & Cluttered: 85.0\%, $\Delta$=10.4 \\
RT-DETR FT & Cluttered: 76.7\%, $\Delta$=22.8 & Cluttered: 81.7\%, $\Delta$=14.3 \\
Mask R-CNN FT & Moderate: 94.3\%, $\Delta$=4.1 & Clean: 96.0\%, $\Delta$=2.3 \\
YOLO-World ZS & Clean: 23.4\%, $\Delta$=24.3 & Cluttered: 16.1\%, $\Delta$=31.0 \\
SAM 3 ZS & Cluttered: 86.1\%, $\Delta$=8.7 & Clean: 87.4\%, $\Delta$=5.7 \\
\bottomrule
\end{tabular}
\vspace{0.4em}
\caption{Clutter difficulty analysis.}
\label{tab:clutter_difficulty}
\end{table}
\vspace{-2.5em}

\textbf{Crowding} is computed as the number of ground-truth bird instances present in the same frame. This factor captures whether models are more likely to miss birds when multiple individuals appear simultaneously in the scene.

\vspace{-0.4em}
\begin{table}[H]
\centering
\scriptsize
\setlength{\tabcolsep}{4pt}
\renewcommand{\arraystretch}{1.12}
\begin{tabular}{lc}
\toprule
\textbf{Model} & \textbf{Crowding} \\
\midrule
YOLO-World FT & \textbf{2--3 objects: 72.9\%, $\Delta$=19.4} \\
RT-DETR FT & 2--3 objects: 87.5\%, $\Delta$=3.1 \\
Mask R-CNN FT & \textbf{2--3 objects: 89.6\%, $\Delta$=7.9} \\
YOLO-World ZS & \textbf{2--3 objects: 10.4\%, $\Delta$=23.3} \\
SAM 3 ZS & 2--3 objects: 87.5\%, $\Delta$=2.9 \\
\bottomrule
\end{tabular}
\vspace{0.4em}
\caption{Crowding difficulty analysis.}
\label{tab:crowding_difficulty}
\end{table}
\vspace{-3em}

\subsection{Multivariate difficulty-factor analysis}

The previous difficulty-factor analysis evaluates each visual factor separately. However, in fixed-camera ecological recordings, difficulty factors may co-occur within the same frame or recording session. For example, a bird instance may appear in a cluttered region, under low local contrast, and within a video with specific illumination or background conditions. To examine these factors jointly, we fitted an exploratory multivariable logistic regression for each model, using object recovery as a two-level response variable and the measured difficulty factors as explanatory predictors~\cite{harris2021binarylogistic}. A ground-truth instance was considered recovered if it matched a prediction under the same greedy IoU-based protocol used in the difficulty-factor analysis. The explanatory variables were the visual and scene-level factors computed in the previous section: relative size, brightness, contrast, clutter, blur/sharpness, and crowding. Before fitting the models, these predictors were z-score standardized to zero mean and unit variance, allowing the resulting coefficients to be interpreted on a common input scale~\cite{gelman2008scaling}. We also included video fixed effects, encoded as session-specific dummy variables, to partially account for unmodeled differences between recording sessions, such as camera viewpoint, background composition, and illumination conditions. This follows the fixed-effects framing for grouped binary-response data, where group-specific intercepts are used to account for unmodeled group-level heterogeneity~\cite{beck2020grouped}. The video coefficients are used only as controls and are therefore not reported in the table.

This analysis is intended as a diagnostic robustness check rather than a definitive explanatory model. The estimated coefficients are associations, not causal effects, and the visual predictors are not independent by design because several difficulty factors can occur together in the same frame. Table~\ref{tab:multivariable_failure_video_fe} reports the logistic-regression coefficients for the z-scored predictors after accounting for the other measured factors and video-level variation. Negative values indicate factors associated with lower recovery probability, whereas positive values indicate factors associated with higher recovery probability within this multivariable specification. Bold values mark the most negative coefficient for each model, highlighting the factor with the strongest negative association with object recovery in that model.

\begin{table}[H]
\centering
\setlength{\tabcolsep}{3pt}
\renewcommand{\arraystretch}{1.15}
\begin{tabular}{lccccccc}
\toprule
\textbf{Model} 
& \textbf{Recall}
& \textbf{Rel. size}
& \textbf{Bright.}
& \textbf{Contrast}
& \textbf{Clutter}
& \textbf{Blur}
& \textbf{Crowding} \\
\midrule
YOLO-World FT 
& 90.55\% & $0.24$  & $\mathbf{-1.37}$ & $-0.63$ & $-1.02$ & $0.39$  & $-0.25$ \\

RT-DETR FT    
& 90.36\% & $-0.04$ & $\mathbf{-2.95}$ & $-2.80$ & $-0.01$ & $0.51$  & $0.14$ \\

Mask R-CNN FT 
& 96.79\% & $-0.00$ & $\mathbf{-5.29}$ & $-3.93$ & $-3.71$ & $0.19$  & $-0.72$ \\

YOLO-World ZS 
& 31.57\% & $0.64$  & $-0.38$ & $-0.19$ & $-0.05$ & $-0.62$ & $\mathbf{-0.83}$ \\

SAM 3 ZS       
& 90.17\% & $0.08$  & $\mathbf{-2.63}$ & $-2.30$ & $-0.03$ & $0.25$  & $0.03$ \\
\bottomrule
\end{tabular}
\vspace{0.4em}
\caption{Exploratory multivariable analysis with video fixed effects. Values are logistic-regression coefficients for z-scored predictors; bold indicates the most negative coefficient for each model.}
\label{tab:multivariable_failure_video_fe}
\end{table}
\vspace{-4em}

\subsection{Discussion}

The diagnostic analyses show that model failures in this benchmark are not explained by a single visual factor. The univariate difficulty analysis indicates that brightness, contrast, blur, clutter, relative size, and crowding can reduce instance recovery. The exploratory multivariable analysis provides a complementary robustness check: once these factors are considered jointly and video-level variation is partially controlled, the most recurrent negative associations are related to appearance, especially brightness and contrast.

This pattern is visible across several models. RT-DETR FT, Mask R-CNN FT, and SAM 3 ZS all show brightness and contrast among their main negative coefficients, suggesting that strong average recall does not imply complete robustness to changes in local or global appearance. YOLO-World FT also shows negative coefficients for brightness, contrast, and clutter, indicating that fine-tuning improves recovery but does not fully remove sensitivity to background complexity and visual appearance.

YOLO-World ZS shows the most distinct failure profile. It has the lowest recall among the evaluated models, and its main negative coefficients are associated with crowding, blur, and brightness. This is consistent with the previous difficulty-factor analysis, where the zero-shot model was especially fragile under degraded local visual quality and multi-object scenes. In contrast, Mask R-CNN FT obtains the highest recall in the diagnostic split, although its coefficients should be interpreted cautiously because the model produces relatively few failures, which can make coefficient estimates less stable.

Overall, the diagnostic results suggest that fixed-camera ecological monitoring should not be treated only as a small-object detection problem. Object scale is part of the difficulty, but the remaining failures also depend on appearance degradation, background complexity, crowding, blur, and session-specific recording conditions.

\section{Conclusion}

We introduced a benchmark for bird detection and segmentation in fixed-camera ecological recordings. The benchmark captures a challenging visual regime in which birds occupy a small fraction of high-resolution frames and appear under clutter, illumination variation, blur, and session-specific viewpoint changes. Our results show that this setting remains challenging across modern vision paradigms. Supervised detectors provide the most consistent overall performance, while zero-shot open-vocabulary models are unreliable and often exhibit unstable precision--recall trade-offs. Fine-tuning substantially improves these models, in some cases making them competitive with supervised approaches, but does not fully eliminate robustness issues. A similar pattern is observed in segmentation, where supervised methods perform strongly, whereas prompt-based segmentation requires task-specific adaptation to be reliable.

Compared with results commonly reported on wildlife and open-vocabulary benchmarks, our benchmark induces a different ranking across model families. On aerial and wildlife detection benchmarks, YOLO-based and transformer-based detectors often outperform two-stage methods such as Faster R-CNN~\cite{mou2023waid,zhou2025uavdetr}. In contrast, Faster R-CNN provides the strongest performance on the primary localization metric in our benchmark, while RT-DETR is stronger mainly in recall and overall balance. A similar shift appears in the open-vocabulary setting: models from the same families achieve substantially stronger results on general-purpose large-vocabulary benchmarks such as LVIS~\cite{gupta2019lvis,li2022glip,yao2025detclipv3,fu2025llmdet}, yet zero-shot OWL-ViT, Grounding DINO, and YOLO-World perform poorly on our dataset~\cite{minderer2022simple,liu2023grounding,cheng2024yoloworld}.

Together, these findings suggest that the main challenge is not category recognition alone, but precise localization of small targets under cluttered backgrounds, changing appearance, and substantial domain shift. The near-collapse of zero-shot open-vocabulary models, together with the sharper trade-offs observed even after adaptation, indicates that fixed-camera ecological recordings capture a failure regime that remains underrepresented in standard detection benchmarks. We hope that this dataset and benchmark support future work on small-object perception, domain adaptation, and robust detection and segmentation methods for ecological applications.

\bibliographystyle{splncs04}
\bibliography{egbib}
\end{document}